\documentclass{article}

\PassOptionsToPackage{numbers, compress}{natbib}
\usepackage[final]{nips_2017}

\usepackage[utf8]{inputenc} 
\usepackage[T1]{fontenc}    
\usepackage{hyperref}       
\usepackage{url}            
\usepackage{booktabs}       
\usepackage{amsfonts}       
\usepackage{nicefrac}       
\usepackage{microtype}      
\usepackage{graphicx} 
\usepackage{epsfig} 
\usepackage{subfigure}
\usepackage{amsmath,multirow,color}
\usepackage{wrapfig}

\newcommand{\lsp}[1]{\large\renewcommand{\baselinestretch}{#1}\normalsize}

\title{Dense Transformer Networks}

\author{
  Jun Li \\
  School of Electrical Engineering\\
  and Computer Science\\
  Washington State University\\
  Pullman, WA 99163\\
  \texttt{jun.li3@wsu.edu}\\
  \And
  Yongjun Chen \\
  School of Electrical Engineering\\
  and Computer Science\\
  Washington State University\\
  Pullman, WA 99163\\
  \texttt{yongjun.chen@wsu.edu}\\
  \And
  Lei Cai \\
  School of Electrical Engineering\\
  and Computer Science\\
  Washington State University\\
  Pullman, WA 99163\\
  \texttt{lei.cai@wsu.edu}\\
  \And
  Ian Davidson \\
  Computer Science Department\\
  University of California, Davis\\
  Davis, CA 95616\\
  \texttt{davidson@cs.ucdavis.edu}\\
  \And
  Shuiwang Ji \\
  School of Electrical Engineering\\
  and Computer Science\\
  Washington State University\\
  Pullman, WA 99163\\
  \texttt{sji@eecs.wsu.edu}\\
}

\begin{document}

\maketitle

\begin{abstract}
The key idea of current deep learning methods for dense prediction
is to apply a model on a regular patch centered on each pixel to
make pixel-wise predictions. These methods are limited in the sense
that the patches are determined by network architecture instead of
learned from data. In this work, we propose the dense transformer
networks, which can learn the shapes and sizes of patches from data.
The dense transformer networks employ an encoder-decoder
architecture, and a pair of dense transformer modules are inserted
into each of the encoder and decoder paths. The novelty of this work
is that we provide technical solutions for learning the shapes and
sizes of patches from data and efficiently restoring the spatial
correspondence required for dense prediction. The proposed dense
transformer modules are differentiable, thus the entire network can
be trained. We apply the proposed networks on natural and biological
image segmentation tasks and show superior performance is achieved
in comparison to baseline methods.
\end{abstract}

\section{Introduction}
\label{introduction}

In recent years, deep convolution neural networks (CNNs) have
achieved promising performance on many artificial intelligence
tasks, including image
recognition~\cite{LeCun:PIEEE,krizhevsky2012imagenet}, object
detection~\cite{sermanetOverFeat,girshick2014rich}, and
segmentation~\cite{farabet2013learning,pinheiro2014recurrent,pinheiro2016learning,chen2014semantic,visin2015reseg}.
Among these tasks, dense prediction tasks take images as inputs and
generate output maps with similar or the same size as the inputs.
For example, in image semantic segmentation, we need to predict a
label for each pixel on the input
images~\cite{long2015fully,noh2015learning}. Other examples include
depth estimation~\cite{laina2016deeper,eigen2014depth}, image
super-resolution~\cite{dong2016image}, and surface normal
prediction~\cite{eigen2015predicting}. These tasks can be generally
considered as image-to-image translation problems in which inputs
are images, and outputs are label maps~\cite{isola2016image}.

Given the success of deep learning methods on image-related
applications, numerous recent attempts have been made to solve dense
prediction problems using CNNs. A central idea of these methods is
to extract a square patch centered on each pixel and apply CNNs on
each of them to compute the label of the center pixel. The
efficiency of these approaches can be improved by using fully
convolutional or encoder-decoder networks. Specifically, fully
convolutional networks~\cite{long2015fully} replace fully connected
layers with convolutional layers, thereby allowing inputs of
arbitrary size during both training and test. In contrast,
deconvolution networks~\cite{noh2015learning} employ an
encoder-decoder architecture. The encoder path extracts high-level
representations using convolutional and pooling layers. The decoder
path uses deconvolutional and up-pooling layers to recovering the
original spatial resolution. In order to transmit information
directly from encoder to decoder, the U-Net~\cite{ronneberger2015u}
adds skip connections~\cite{he2016deep} between the corresponding
encoder and decoder layers. A common property of all these methods
is that the label of any pixel is determined by a regular (usually
square) patch centered on that pixel. Although these methods have
achieved considerable practical success, there are limitations
inherent in them. For example, once the network architecture is
determined, the patches used to predict the label of each pixel is
completely determined, and they are commonly of the same size for
all pixels. In addition, the patches are
usually of a regular shape, e.g., squares.

In this work, we propose the dense transformer networks to address
these limitations. Our method follows the encoder-decoder
architecture in which the encoder converts input images into
high-level representations, and the decoder tries to make pixel-wise
predictions by recovering the original spatial resolution. Under
this framework, the label of each pixel is also determined by a
local patch on the input. Our method allows the size and shape of
every patch to be adaptive and data-dependent. In order to achieve
this goal, we propose to insert a spatial transformer
layer~\cite{jaderberg2015spatial} in the encoder part of our
network. We propose to use nonlinear transformations, such as these
based on thin-plate
splines~\cite{shi2016robust,bookstein1989principal}. The nonlinear
spatial transformer layer transforms the feature maps into a
different space. Therefore, performing regular convolution and
pooling operations in this space corresponds to performing these
operations on irregular patches of different sizes in the original
space. Since the nonlinear spatial transformations are learned
automatically from data, this corresponds to learning the size and
shape of each patch to be used as inputs for convolution and pooling
operations.

There has been prior work on allowing spatial transformations or deformations in
deep networks~\cite{jaderberg2015spatial,dai2017deformable}, but
they do not address the spatial correspondence problem, which is critical in dense prediction tasks. The difficulty in applying
spatial transformations to dense prediction tasks lies in that the
spatial correspondence between input images and output label maps
needs to be preserved. A key innovation of this work is that we
provide a new technical solution that not only allows data-dependent
learning of patches but also enables the preservation of spatial
correspondence. Specifically, although the patches used to predict
pixel labels could be of different sizes and shapes, we expect the
patches to be in the spatial vicinity of pixels whose labels are to
be predicted. By applying the nonlinear spatial transformer layers
in the encoder path as described above, the spatial locations of
units on the intermediate feature maps after the spatial
transformation layer may not be preserved. Thus a reverse
transformation is required to restore the spatial correspondence.

In order to restore the spatial correspondence between inputs and
outputs, we propose to add a corresponding decoder layer. A
technical challenge in developing the decoder layer is that we need
to map values of units arranged on input regular grid to another set
of units arranged on output grid, while the nonlinear transformation
could map input units to arbitrary locations on the output map. We
develop a interpolation method to address this challenge.
Altogether, our work results in the dense transformer networks,
which allow the prediction of each pixel to adaptively choose the
input patch in a data-dependent manner. The dense transformer
networks can be trained end-to-end, and gradients can be
back-propagated through both the encoder and decoder layers.
Experimental results on natural and biological images demonstrate
the effectiveness of the proposed dense transformer networks.

\section{Spatial Transformer Networks Based on Thin-Plate Spline}
\label{sec:STN}

Spatial transformer networks \cite{jaderberg2015spatial} are deep
models containing spatial transformer layers. These layers
explicitly compute a spatial transformation of the input feature
maps. They can be inserted into convolutional neural networks to
perform explicit spatial transformations. The spatial transformer
layers consist of three components; namely, the localization
network, grid generator and sampler.

The localization network takes a set of feature maps as input and
generates parameters to control the transformation. If there are
multiple feature maps, the same transformation is applied to all of
them. The grid generator constructs transformation mapping between
input and output grids based on parameters computed from the
localization network. The sampler computes output feature maps based
on input feature maps and the output of grid generator. The spatial
transformer layers are generic and different types of
transformations, e.g., affine transformation, projective
transformation, and thin-plate spline (TPS), can be used. Our
proposed work is based on the TPS transformation, and it is not
described in detail in the original
paper~\cite{jaderberg2015spatial}. Thus we provide more details
below.

\subsection{Localization Network}

When there are multiple feature maps, the same transformation is
applied to all of them. Thus, we assume there is only one input
feature map below. The TPS transformation is determined by $2K$
fiducial points among which $K$ points lie on the input feature map
and the other $K$ points lie on the output feature map. On the
output feature map, the $K$ fiducial points, whose coordinates are
denoted as $\tilde F=[\tilde f_1,\tilde f_2,\cdots,\tilde f_K] \in
\mathbb{R}^{2 \times K} $, are evenly distributed on a fixed regular
grid, where $\tilde f_i=[\tilde x_i, \tilde y_i]^T$ denotes the
coordinates of the $i$th point. The localization network is used to
learn the $K$ fiducial points $F=[f_{1},f_{2},\cdots,f_{K}] \in
\mathbb{R}^{2 \times K}$ on the input feature map. Specifically, the
localization network, denoted as $f_{loc}(\cdot)$, takes the input
feature maps $U \in \mathbb{R}^{H \times W \times C}$ as input,
where $H$, $W$ and $C$ are the height, width and number of channels
of input feature maps, and generates the normalized coordinates $F$
as the output as $F=f_{loc}(U)$.

A cascade of convolutional, pooling and fully-connected layers
is used to implement $f_{loc}(\cdot)$.
The output of the final fully-connected layer is the coordinates $F$
on the input feature map. Therefore, the number of output units of
the localization network is $2K$. In order to ensure that the
outputs are normalized between $-1$ and $1$, the activation function
$\tanh(\cdot)$ is used in the fully-connected layer. Since the
localization network is differentiable, the $K$ fiducial points can
be learned from data using error back-propagation.

\subsection{Grid Generator}

For each unit lying on a regular grid on the output feature map, the
grid generator computes the coordinate of the corresponding unit on
the input feature map. This correspondence is determined by the
coordinates of the fiducial points $F$ and $\tilde F$. Given the
evenly distributed $K$ points $\tilde F=[\tilde f_1,\tilde
f_2,\cdots,\tilde f_K]$ on the output feature map and the $K$
fiducial points $F=[f_{1},f_{2},\cdots,f_{K}]$ generated by the
localization network, the transformation matrix $T$ in TPS can be
expressed as follows:
\begin{equation}\label{eq:transformation matrix}
T = \left({\Delta_{\tilde F}^{-1}}\times\left[\begin{array}{cccc}
    {F^{T}}\\
    {{\bf 0}}^{3 \times 2}
\end{array}\right]\right)^{T} \in \mathbb{R}^{2 \times (K+3)},
\end{equation}
where $\Delta_{\tilde F} \in \mathbb{R}^{(K+3) \times (K+3)}$ is a
matrix determined only by $\tilde F$ as
\begin{equation}\label{eq: F hat}
{\Delta_{\tilde F}} = \left[\begin{array}{cccc}
    \mathbf{1}^{K \times 1}  &     \tilde F^T    & {R}\\
    \mathbf{0}^{1\times 1} &    \mathbf{0}^{1\times 2}    & \mathbf{1}^{1 \times K}\\
    \mathbf{0}^{2\times 1} &    \mathbf{0}^{2\times 2}    & \tilde F
\end{array}\right]\in \mathbb{R}^{(K+3) \times (K+3)},
\end{equation}
where $R\in\mathbb{R}^{K\times K}$, and its elements are defined as
$r_{i,j} = d_{i,j}^2 \ln{d_{i,j}^2}$, and $d_{i,j}$ denotes the
Euclidean distance between ${\tilde f_i}$ and $\tilde {f_j}$.

Through the mapping, each unit $(\tilde x_i, \tilde y_i)$ on the
output feature map corresponds to unit $(x_i, y_i)$ on the input
feature map. To achieve this mapping, we represent the units on the
regular output grid by $\{{\tilde p_i}\}_{i=1}^{\tilde H \times
\tilde W}$, where ${\tilde p_i} = [\tilde x_i, \tilde y_i]^{T}$ is
the $(x, y)$-coordinates of the $i$th unit on output grid, and
$\tilde H$ and $\tilde W$ are the height and width of output feature
maps. Note that the fiducial points $\{\tilde f_i\}_{i=1}^K$ are a
subset of the points $\{{\tilde p_i}\}_{i=1}^{\tilde H \times \tilde
W}$, which are the set of all points on the regular output grid.

To apply the transformation, each point $\tilde p_i$ is first
extended from $\mathbb{R}^2$ space to $\mathbb{R}^{K+3}$ space as
$
{\tilde q_{i}} = [1, \tilde x_i, \tilde y_i, s_{i,1},
s_{i,2},\cdots,s_{i,K}]^{T}\in\mathbb{R}^{K+3},
$
where $s_{i,j} = e_{i,j}^2 \ln{e_{i,j}^2}$, and $e_{i,j}$ is the
Euclidean distance between $\tilde p_i$ and ${\tilde f_j}$. Then the
transformation can be expressed as
\begin{equation}\label{eq: project eq}
 {p_{i}} = {T\tilde q_{i}},
\end{equation}
where $T$ is defined in Eq.~(\ref{eq:transformation matrix}). By
this transformation, each coordinate $(\tilde x_i, \tilde y_i)$ on
the output feature map corresponds to a coordinate $(x_i, y_i)$ on
the input feature map. Note that the transformation $T$ is defined
so that the points $\tilde F$ map to points $F$.

\subsection{Sampler}

The sampler generates output feature maps based on input feature
maps and the outputs of grid generator. Each unit $\tilde p_i$ on
the output feature map corresponds to a unit $p_i$ on the input
feature map as computed by Eq.~(\ref{eq: project eq}). However, the
coordinates $p_{i}= (x_i, y_i)^T$ computed by Eq.~(\ref{eq: project
eq}) may not lie exactly on the input regular grid. In these cases,
the output values need to be interpolated from input values lying on
regular grid. For example, a bilinear sampling method can be used to
achieve this. Specifically, given an input feature map
$U\in\mathbb{R}^{H\times W}$, the output feature map
$V\in\mathbb{R}^{\tilde H\times \tilde W}$ can be obtained as
\begin{equation}\label{eq:bilinear:sample}
\hspace{-0.16cm}{V_{i}\hspace{-0.07cm}=\hspace{-0.07cm}\sum_{n=1}^{H}\hspace{-0.07cm}\sum_{m=1}^{W}\hspace{-0.07cm}U_{nm}\max(0,1\hspace{-0.07cm}-|x_{i}\hspace{-0.07cm}-m|)\max(0,1\hspace{-0.07cm}-|y_{i}\hspace{-0.07cm}-n|)}
\end{equation}
for $i = 1, 2,\cdots,\tilde H\times\tilde W$, where $V_{i}$ is the
value of pixel $i$, $U_{nm}$ is the value at $(n,m)$ on the input
feature map, $p_i=(x_{i}, y_{i})^T$, and $p_i$ is computed from
Eq.~(\ref{eq: project eq}). By using the transformations, the
spatial transformer networks have been shown to be invariant to some
transformations on the inputs. Other recent studies have also
attempted to make CNNs to be invariant to various
transformations~\cite{jia2016dynamic,henriques2016warped,cohen2016group,dieleman2016exploiting}.

\section{Dense Transformer Networks}
\label{sec:DTN}

The central idea of CNN-based method for dense prediction is to
extract a regular patch centered on each pixel and apply CNNs to
compute the label of that pixel. A common property of these methods
is that the label of each pixel is determined by a regular
(typically square) patch centered on that pixel. Although these
methods have been shown to be effective on dense prediction
problems, they lack the ability to learn the sizes and shapes of
patches in a data-dependent manner. For a given network, the size of
patches used to predict the labels of each center pixel is
determined by the network architecture. Although multi-scale
networks have been proposed to allow patches of different sizes to
be combined~\cite{farabet2013learning}, the patch sizes are again
determined by network architectures. In addition, the shapes of
patches used in CNNs are invariably regular, such as squares.
Ideally, the shapes of patches may depend on local image statistics
around that pixel and thus should be learned from data. In this
work, we propose the dense transformer networks to enable the
learning of patch size and shape for each pixel.

\begin{figure*}[t]
\vskip 0.2in
\begin{center}
{\includegraphics[width=1.0\textwidth]{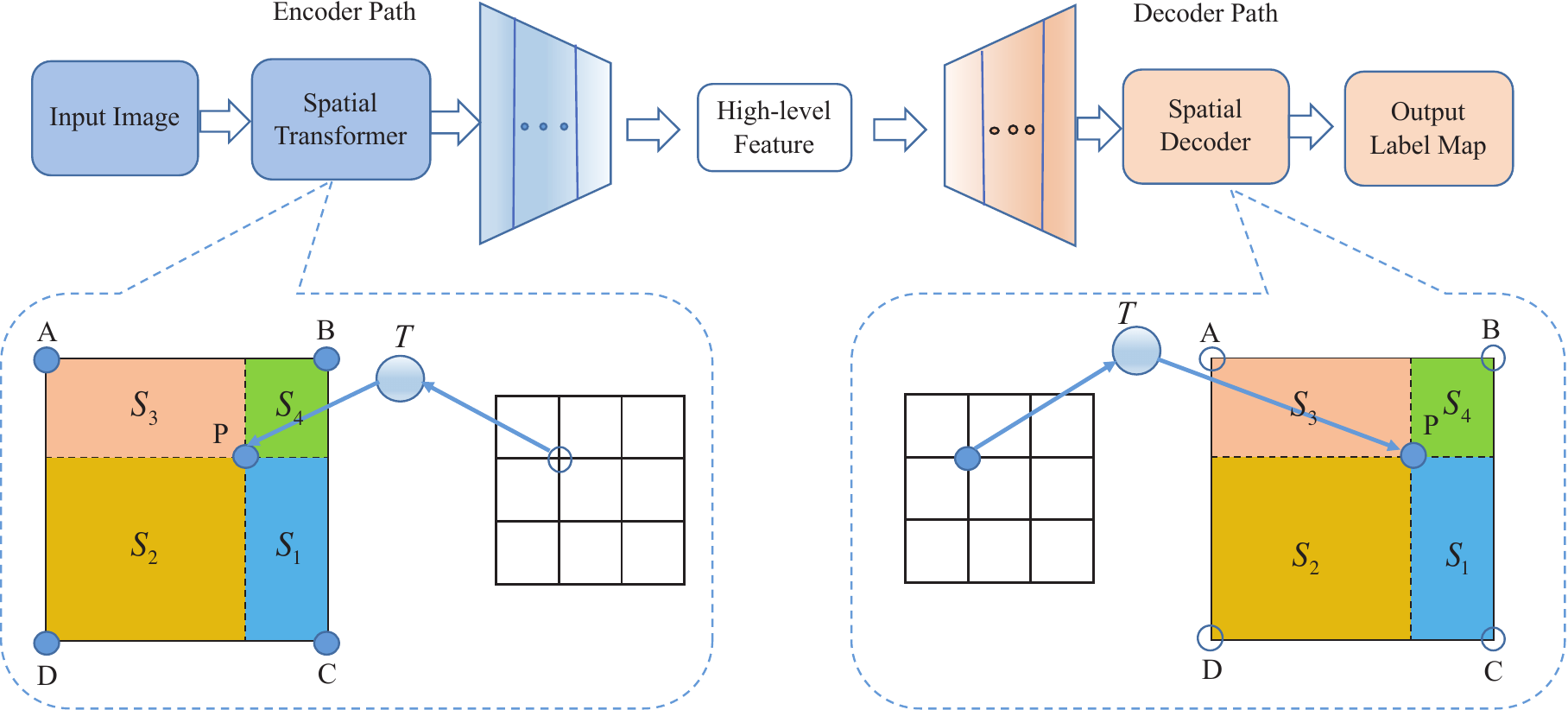}}\vspace{-0.3cm}
\caption{The proposed dense transformer networks. A pair of dense
transformer modules are inserted into each of the encoder and
decoder paths. In the spatial transformer module, values at points
$A$, $B$, $C$, and $D$ are given from the previous layer, and we
need to estimate value for point $P$. In contrast, in the decoder
layer, value at point $P$ is given from the previous layer, and we
need to estimate values for points $A$, $B$, $C$, and $D$.}
\label{fig:DTN}
\end{center}
\vskip 0.2in
\end{figure*}

\subsection{An Encoder-Decoder Architecture}

In order to address the above limitations, we propose to develop a
dense transformer network model. Our model employs an
encoder-decoder architecture in which the encoder path extracts
high-level representations using convolutional and pooling layers
and the decoder path uses deconvolution and un-pooling to recover
the original spatial
resolution~\cite{noh2015learning,ronneberger2015u,badrinarayanan2015segnet,newell2016stacked}.
To enable the learning of size and shape of each patch automatically
from data, we propose to insert a spatial transformer module in the
encoder path in our network. As has been discussed above, the
spatial transformer module transforms the feature maps into a
different space using nonlinear transformations. Applying
convolution and pooling operations on regular patches in the
transformed space is equivalent to operating on irregular patches of
different sizes in the original space. Since the spatial transformer
module is differentiable, its parameters can be learned with error
back-propagation algorithms. This is equivalent to learning the size
and shape of each patch from data.

Although the patches used to predict pixel labels could be of
different sizes and shapes, we expect the patches to include the
pixel in question at least. That is, the patches should be in the
spatial vicinity of pixels whose labels are to be predicted. By
using the nonlinear spatial transformer layer in encoder path, the
spatial locations of units on the intermediate feature maps could
have been changed. That is, due to this nonlinear spatial
transformation, the spatial correspondence between input images and
output label maps is not retained in the feature maps after the
spatial transformer layer. In order to restore this spatial
correspondence, we propose to add a corresponding decoder layer,
known as the dense transformer decoder layer. This decoder layer
transforms the intermediate feature maps back to the original input
space, thereby re-establishing the input-output spatial
correspondence.

The spatial transformer module can be inserted after any layer in
the encoder path while the dense transform decoder module should be
inserted into the corresponding location in decoder path. In our
framework, the spatial transformer module is required to not only
output the transformed feature maps, but also the transformation
itself that captures the spatial correspondence between input and
output feature maps. This information will be used to restore the
spatial correspondence in the decoder module. Note that in the
spatial transformer encoder module, the transformation is computed
in the backward direction, i.e., from output to input feature maps
(Figure~\ref{fig:DTN}). In contrast, the dense transformer decoder
module uses a forward direction instead; that is, a mapping from
input to output feature maps. This encoder-decoder pair can be
implemented efficiently by sharing the transformation parameters in
these two modules.

A technical challenge in developing the dense transformer decoder
layer is that we need to map values of units arranged on input
regular grid to another set of units arranged on regular output
grid, while the decoder could map to units at arbitrary locations on
the output map. That is, while we need to compute the values of
units lying on regular output grid from values of units lying on
regular input grid, the mapping itself could map an input unit to an
arbitrary location on the output feature map, i.e., not necessarily
to a unit lying exactly on the output grid. To address this
challenge, we develop a sampler method for performing interpolation.
We show that the proposed samplers are differentiable, thus
gradients can be propagated through these modules. This makes the
entire dense transformer networks fully trainable. Formally, assume
that the encoder and decoder layers are inserted after the $i$-th
and $j$-th layers, respectively, then we have the following
relationships:
\begin{equation}
U^{i+1}(p) = \mbox{Sampling}\{U^i(Tp)\},\,\,\,\,\, U^{j+1}(Tp) =
U^j(p),\,\,\,\,\,U^{j+1}(p) =\mbox{Sampling}\{U^{j+1}(Tp)\},
\end{equation}
where $U^i$ is the feature map of the $i$-th layer, $p$ is the
coordinate of a point, $T$ is the transformation defined in
Eq.~(\ref{eq:transformation matrix}), which maps from the
coordinates of the $(i+1)$-th layer to the $i$-th layer,
$\mbox{Sampling}(\cdot)$ denotes the sampler function.

From a geometric perspective, a value associated with an estimated
point in bilinear interpolation in Eq.~(\ref{eq:bilinear:sample})
can be interpreted as a linear combination of values at four
neighboring grid points. The weights for linear combination are
areas of rectangles determined by the estimated points and four
neighboring grid points. For example, in Figure~\ref{fig:DTN}, when
a point is mapped to $P$ on input grid, the contributions of points
$A$, $B$, $C$, and $D$ to the estimated point $P$ is determined by
the areas of the rectangles $S_1$, $S_2$, $S_3$, and $S_4$. However,
the interpolation problem needs to be solved in the dense
transformer decoder layer is different with the one in the spatial
transformer encoder layer, as illustrated in Figure~\ref{fig:DTN}.
Specifically, in the encoder layer, the points $A$, $B$, $C$, and
$D$ are associated with values computed from the previous layer, and
the interpolation problem needs to compute a value for $P$ to be
propagated to the next layer. In contrast, in the decoder layer, the
point $P$ is associated with a value computed from the previous
layer, and the interpolation problem needs to compute values for
$A$, $B$, $C$, and $D$. Due to the different natures of the
interpolation problems need to be solved in the encoder and decoder
modules, we propose a new sampler that can efficiently interpolate
over decimal points in the following section.

\subsection{Decoder Sampler}

In the decoder sampler, we need to estimate values of regular grid
points based on those from arbitrary decimal points, i.e., those
that do not lie on the regular grid. For example, in Figure
\ref{fig:DTN}, the value at point $P$ is given from the previous
layer. After the TPS transformation in Eq. (\ref{eq: project eq}),
it may be mapped to an arbitrary point. Therefore, the values of
grid points $A$, $B$, $C$, and $D$ need to be computed based on
values from a set of arbitrary points. If we compute the values from
surrounding points as in the encoder layer, we might have to deal
with a complex interpolation problem over irregular quadrilaterals.
Those complex interpolation methods may yield more accurate results,
but we prefer a simpler and more efficient method in this work.
Specifically, we propose a new sampling method, which distributes
the value of $P$ to the points $A$, $B$, $C$, and $D$ in an
intuitive manner. Geometrically, the weights associated with points
$A$, $B$, $C$, and $D$ are the area of the rectangles $S_1$, $S_2$,
$S_3$, and $S_4$, respectively (Figure \ref{fig:DTN}). In
particular, given an input feature map $V\in\mathbb{R}^{\tilde
H\times \tilde W}$, the output feature map $U\in\mathbb{R}^{H\times
W}$ can be obtained as
\begin{eqnarray}
S_{nm}&=& \sum_{i=1}^{\tilde H \times \tilde W}
\max(0,1-|x_{i}-m|)\max(0,1-|y_{i}-n|),\label{eq:bilinear scattering:weight}\\
U_{nm}&=&\frac{1}{S_{nm}} \sum_{i=1}^{\tilde H \times \tilde W}V_{i}
\max(0,1-|x_{i}-m|)\max(0,1-|y_{i}-n|),\label{eq:bilinear
scattering:sample}
\end{eqnarray}
where $V_{i}$ is the value of pixel $i$, $p_i=(x_{i}, y_{i})^T$ is
transformed by the shared transformation $T$ in Eq.
(\ref{eq:transformation matrix}), $U_{nm}$ is the value at the
$(n,m)$-th location on the output feature map, $S_{nm}$ is a
normalization term that is used to eliminate the effect that
different grid points may receive values from different numbers of
arbitrary points, and $n = 1, 2,\cdots,N,$ $m=1, 2, \cdots,M$.

In order to allow the backpropagation of errors, we define the
gradient with respect to $U_{nm}$ as $dU_{nm}$. Then the gradient
with respect to $V_{nm}$ and $x_{i}$ can be derived as follows:
\begin{eqnarray}
\hspace{-0.5cm}dV_{i} \hspace{-0.3cm}&=&
\hspace{-0.3cm}\sum_{n=1}^{H}\sum_{m=1}^{W} \frac{1}{S_{nm}}dU_{nm}
\max(0,1-|x_{i}-m|)\max(0,1-|y_{i}-n|),\label{eq:bilinear scattering
backprop dU}\\
\hspace{-0.5cm}dS_{nm} \hspace{-0.3cm}&=& \hspace{-0.3cm}
\frac{-dU_{nm}}{S_{nm}^2} \sum_{i=1}^{\tilde H \times \tilde
W}V_{i} \max(0,1-|x_{i}-m|)\max(0,1-|y_{i}-n|),\label{eq:bilinear
scattering backprop dS}\\
\hspace{-0.5cm}dx_{i}\hspace{-0.3cm}&=&
\hspace{-0.3cm}\sum_{n=1}^{H}\sum_{m=1}^{W}
\left\{\frac{dU_{nm}}{S_{nm}}V_{i} +dS_{nm}\right\}
\max(0,1-|y_{i}-n|)\times \left\{
\begin{array}{rcl}
0        & \mbox{if}   & |m\hspace{-0.07cm}-x_{i}|\geq 1 \\
1        & \mbox{if}   & m \geq x_{i}\\
-1       & \mbox{if}   & m \le x_{i}
\end{array} \right..\label{eq:bilinear scattering backprop dx}
\end{eqnarray}
A similar gradient can be derived for $dy_{i}$. This provides us
with a differentiable sampling mechanism, which enables the
gradients flow back to both the input feature map and the sampling
layers.

\section{Experimental Evaluation}

We evaluate the proposed methods on two image segmentation tasks.
The U-Net~\cite{ronneberger2015u} is adopted as our base model in
both tasks, as it has achieved state-of-the-art performance on image
segmentation tasks. Specifically, U-Net adds residual connections
between the encoder path and decoder path to incorporate both
low-level and high-level features. Other methods like
SegNet~\cite{badrinarayanan2015segnet}, deconvolutional
networks~\cite{zeiler2010deconvolutional} and
FCN~\cite{long2015fully} mainly differ from U-Net in the up-sampling
method and do not use residual connections. Experiments in prior
work show that residual connections are important while different
up-sampling methods lead to similar results. The network consists of
5 layers in the encoder path and another corresponding 5 layers in
the decoder path. We use 3$\times$3 kernels and one pixel padding to
retain the size of feature maps at each level.

In order to efficiently implement the transformations, we insert the
spatial encoder layer and dense transformer decoder layer into
corresponding positions at the same level. Specifically, the layers
are applied to the 4th layer, and their performance is compared to
the basic U-Net model without spatial transformations. As for the
transformation layers, we use 16 fiducial points that are evenly
distributed on the output feature maps. In the dense transformer
decoder layer, if there are pixels that are not selected on the
output feature map, we apply an interpolation strategy over its
neighboring pixels on previous feature maps to produce smooth
results.

\subsection{Natural Image Semantic Segmentation}

\begin{figure}[!t]
\vskip 0.2in
\begin{center}
\includegraphics[width=\textwidth,height=5.5cm]{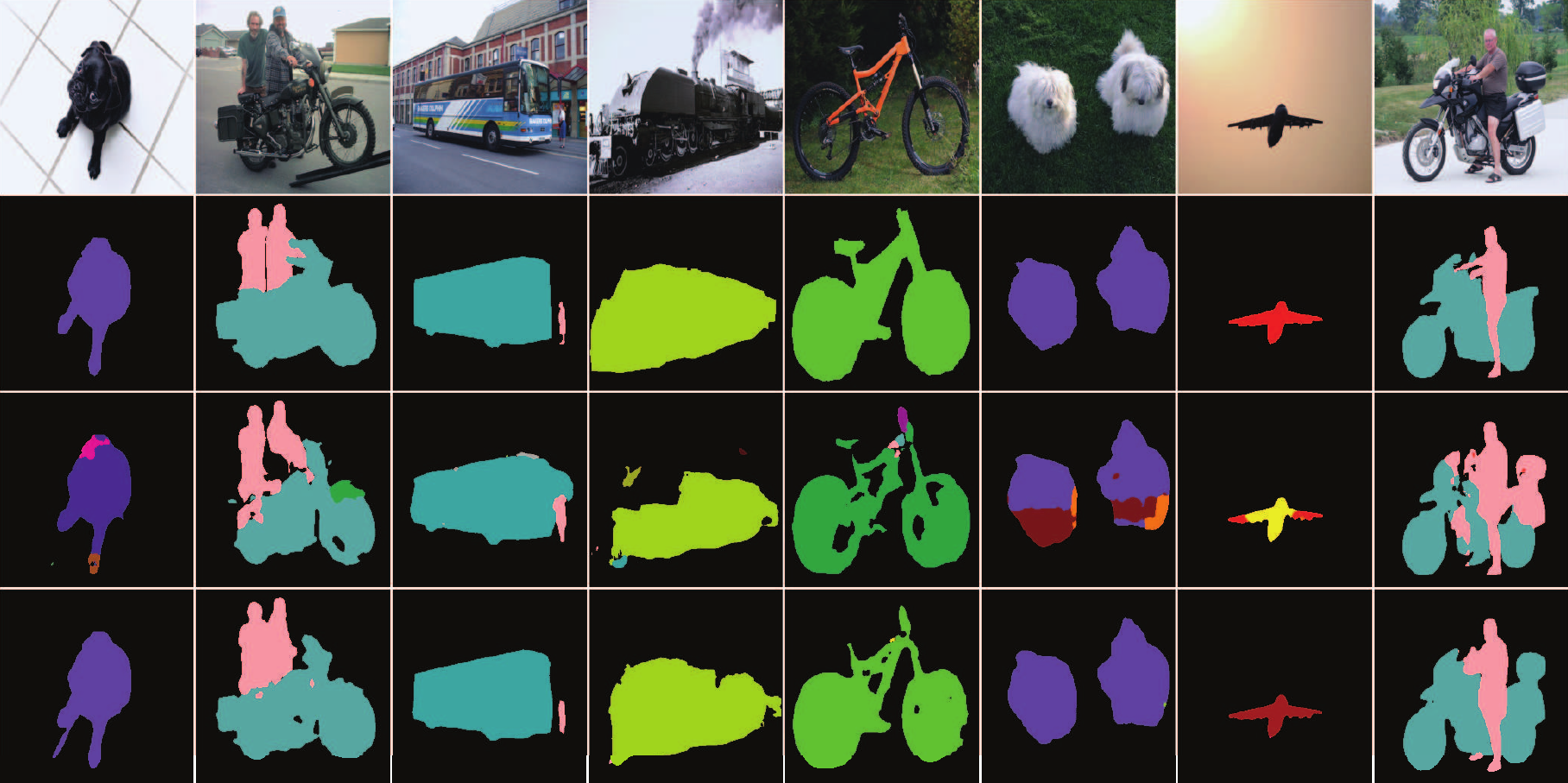}
\end{center}\vspace{-0.3cm}
\caption{Sample segmentation results on the PASCAL 2012 segmentation
data set. The first and second rows are the original images and the
corresponding ground truth, respectively. The third and fourth rows
are the segmentation results of U-Net and DTN, respectively.}
\label{fig:Pascal_result} \vskip 0.2in
\end{figure}

We use the PASCAL 2012 segmentation data
set~\cite{everingham2010pascal} to evaluate the proposed methods on
natural image semantic segmentation task. In this task, we predict
one label out of a total of 21 classes for each pixel. To avoid the
inconvenience of different sizes of images, we resize all the images
to 256$\times$256.
Multiple performance metrics, including loss, accuracy, and
mean-IOU, are used to measure the segmentation performance, and the
results are reported in Table \ref{exp-Pascal-table}. We can observe
that the proposed DTN model achieves higher performance than the
baseline U-Net model. Especially, it improves the mean-IOU from
0.4145 to 0.5297. Some example results along with the raw images and
ground truth label maps are given in Figure~\ref{fig:Pascal_result}.
These results demonstrate that the proposed DTN model can boost the
segmentation performance dramatically.

\begin{table*}[t]
\caption{Comparison of segmentation performance between the U-Net
and the proposed DTN on the PASCAL 2012 segmentation data set. Three
different performance measures are used here. An arrow is attached
to each measure so that $\uparrow$ denotes higher values indicate
better performance, and $\downarrow$ denotes lower values indicate
better performance.} \label{exp-Pascal-table} \vskip 0.15in
\begin{center}
\begin{small}
\begin{sc}
\begin{tabular}{lccccccc}
\hline
Data Set & Model & Loss$\downarrow$  & Accuracy$\uparrow$ & Mean-IOU$\uparrow$\\
\hline
\multirow{2}{*}{PASCAL} & U-Net & 0.9396 & 0.8117 & 0.4145\\
  &DTN & \textbf{0.7909} & \textbf{0.8367} & \textbf{0.5297}\\
\hline
\end{tabular}
\end{sc}
\end{small}
\end{center}
\vskip -0.1in
\end{table*}

\subsection{Brain Electron Microscopy Image Segmentation}

We evaluate the proposed methods on brain electron microscopy (EM)
image segmentation task~\cite{lee2015recursive,ciresan2012deep}, in
which the ultimate goal is to reconstruct neurons at the micro-scale
level. A critical step in neuron reconstruction is to segment the EM
images. We use data set from the 3D Segmentation of Neurites in EM
Images (SNEMI3D, http://brainiac2.mit.edu/SNEMI3D/). The SNEMI3D
data set consists of 100 1024$\times$1024 EM image slices. Since we
perform 2D transformations in this work, each image slice is
segmented separately in our experiments.
\begin{wrapfigure}{r}{0.5\textwidth}
  \vspace{-20pt}
  \begin{center}
    \includegraphics[width=0.48\textwidth]{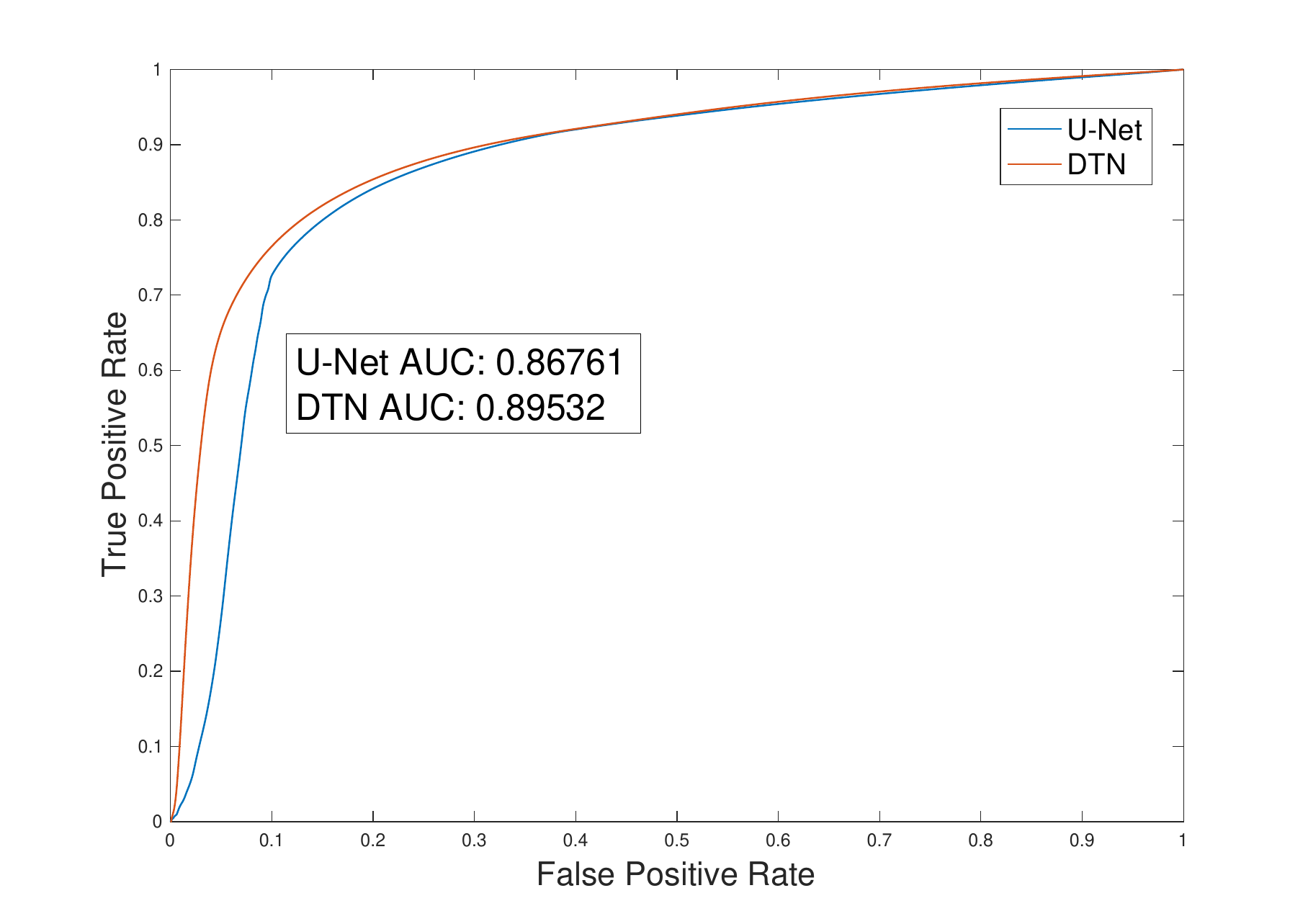}
  \end{center}
  \vspace{-10pt}
  \caption{Comparison of the ROC curves of the U-Net and the proposed DTN model on the SNEMI3D data set.}
  \label{fig:ROCS}
  \vspace{-10pt}
\end{wrapfigure}
The task is to predict each
pixel as either a boundary (denoted as 1) or a non-boundary pixel
(denoted as 0).

\begin{figure}[!t]
\vskip 0.2in
\begin{center}
$\begin{array}{cccccc}
\includegraphics[width=1.2in]{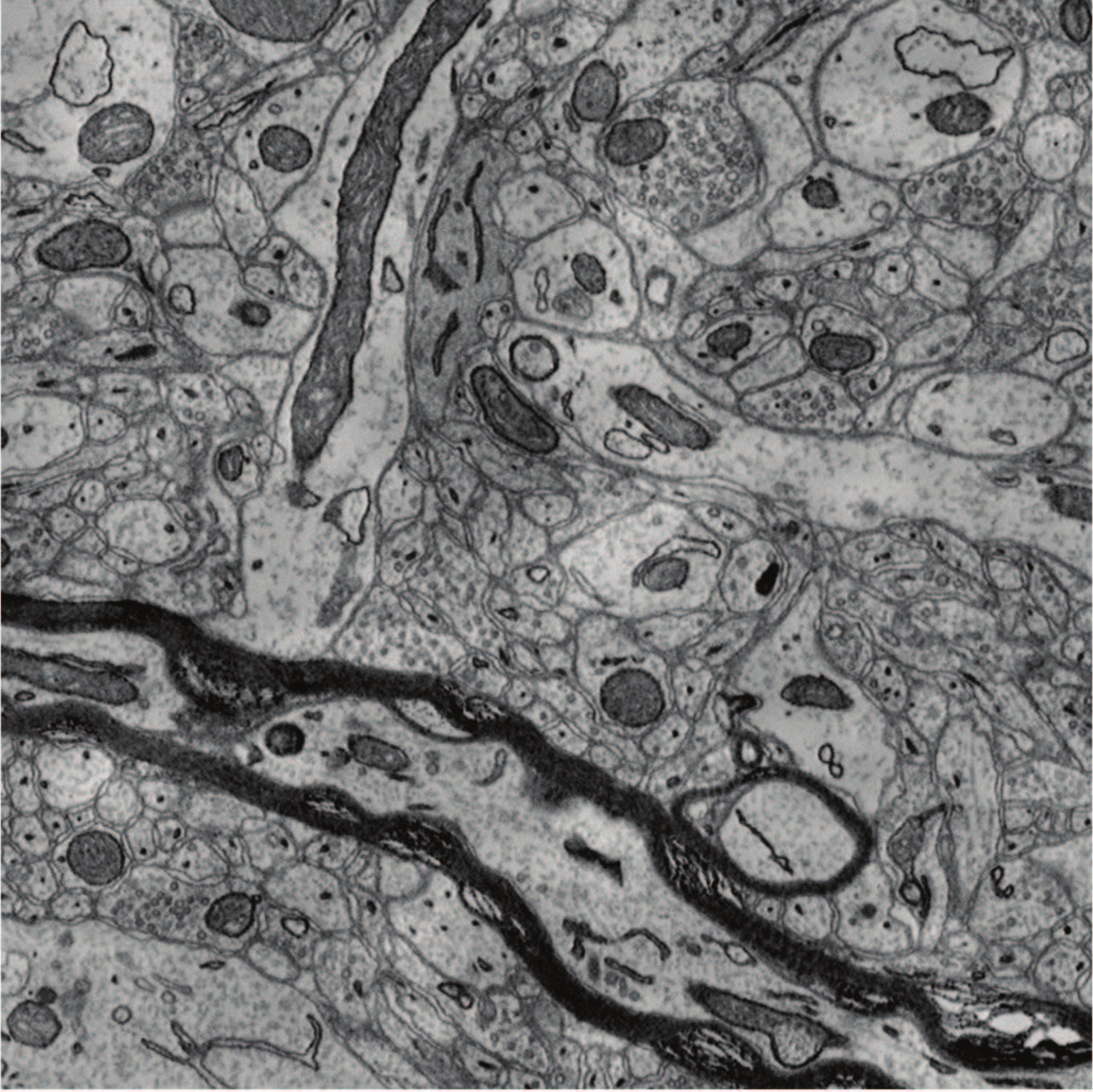}&
\includegraphics[width=1.2in]{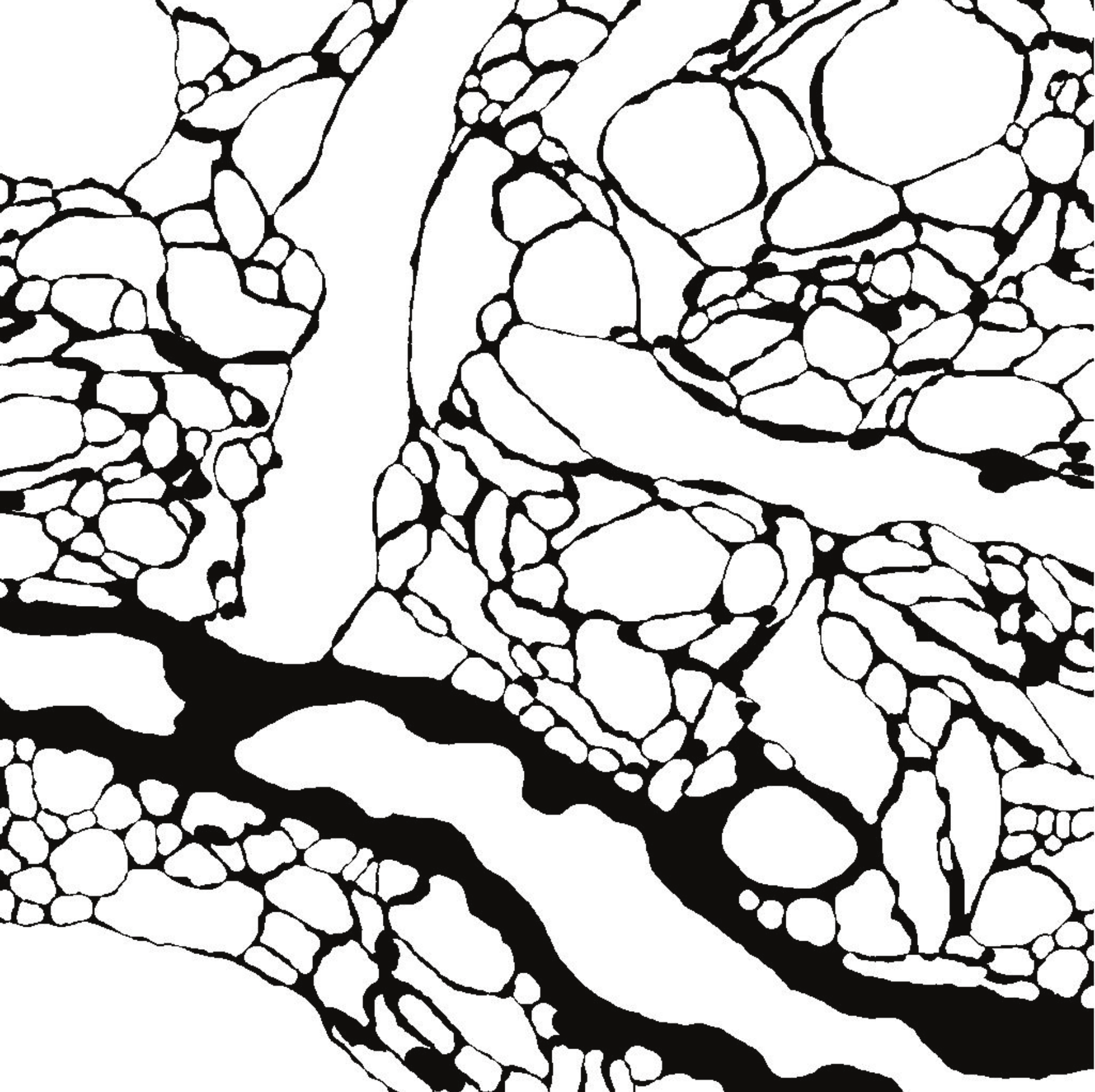}&
\includegraphics[width=1.2in]{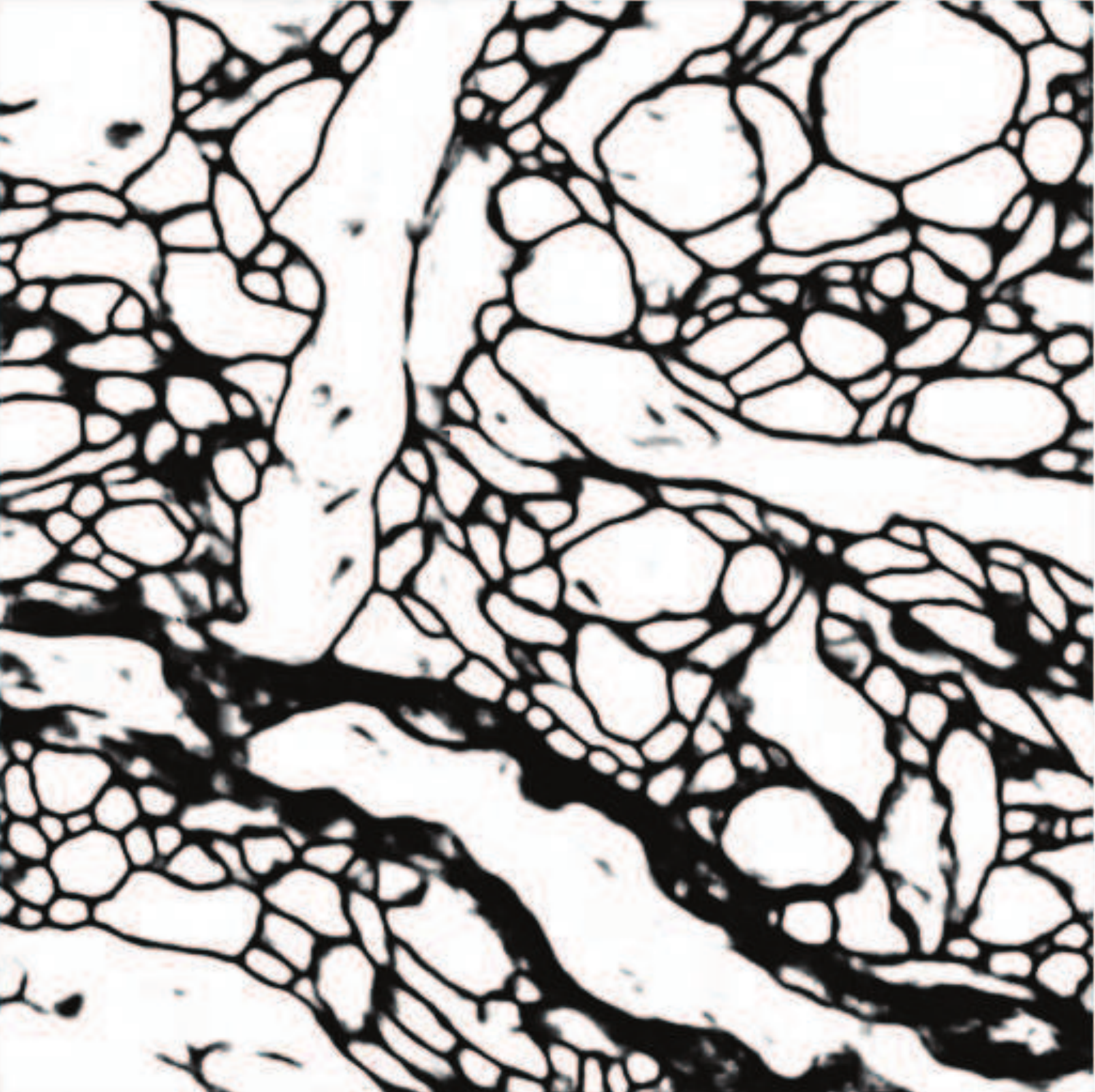}&
\includegraphics[width=1.2in]{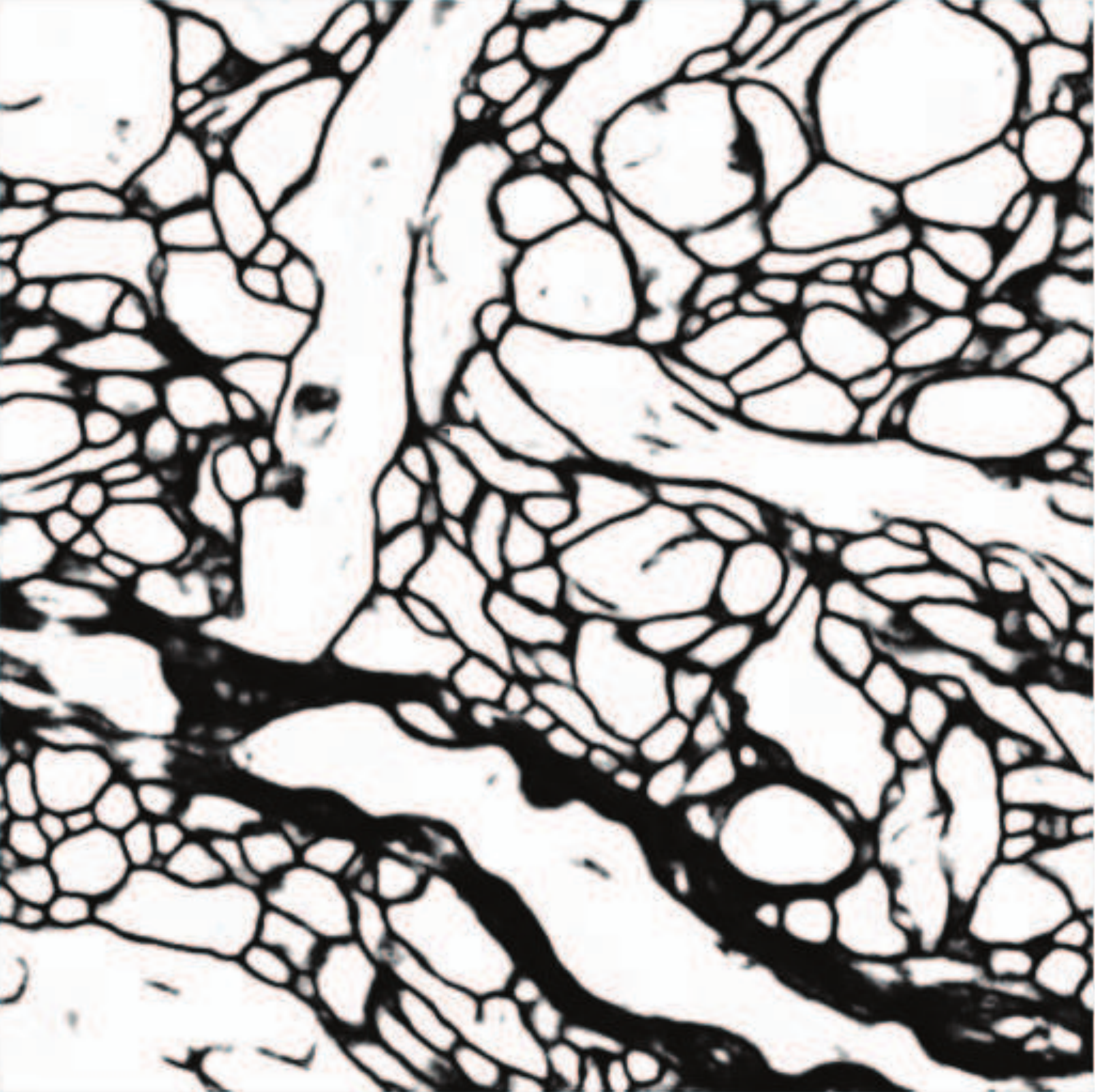}\\
\mbox{Raw image}& \mbox{Ground truth}& \mbox{U-Net output}&
\mbox{DTN output}
\end{array}$
\end{center}\vspace{-0.3cm}
\caption{Example results generated by the U-Net and the proposed DTN
models for the SNEMI3D data set.} \label{fig:EM_result} \vskip 0.2in
\end{figure}

Our model can process images of arbitrary size. However, training on
whole images may incur excessive memory requirement. In order to
accelerate training, we randomly pick 224$\times$224 patches from
the original images and use it to train the networks.
The experimental results in terms of ROC curves are provided in
Figure~\ref{fig:ROCS}. We can observe that the proposed DTN model
achieves higher performance than the baseline U-Net model, improving
AUC from 0.8676 to 0.8953. These results demonstrate that the
proposed DTN model improves upon the baseline U-Net model, and the
use of the dense transformer encoder and decoder modules in the
U-Net architecture results in improved performance. Some example
results along with the raw images and ground truth label maps are
given in Figure~\ref{fig:EM_result}.

\subsection{Timing Comparison}
Table~\ref{exp-time-table} shows the comparison of training and
prediction time between the U-Net model and the proposed DTN model
on the two data sets. We can see that adding DTN layers leads to
only slight increase in training and prediction time. Since the
PASCAL data set is more complex than the SNEMEI3D data set, we use
more channels when building the network of natural image
segmentation task. That causes the increase of training and
prediction time on the PASCAL data set as compared to SNEMEI3D.

\begin{table*}[t]
\caption{Training and prediction time on the two data sets using a
Tesla K40 GPU. We compare the training time of 10,000 iterations and
prediction time of 2019 (PASCAL) and 40 (SNEMI3D) images for the
base U-Net model and the DTN.} \label{exp-time-table} \vskip 0.15in
\begin{center}
\begin{small}
\begin{sc}
\begin{tabular}{lccccccc}
\hline
Data Set  & Model & Training Time  & Prediction Time\\
\hline
\multirow{2}{*}{PASCAL} & U-Net & 378m57s & 14m06s \\
  &DTN & 402m07s & 15m50s \\
\hline
\multirow{2}{*}{SNEMI3D} & U-Net & 14m18s & 3m31s \\
  &DTN & 15m41s & 4m02s\\
\hline
\end{tabular}
\end{sc}
\end{small}
\end{center}
\vskip -0.1in
\end{table*}

\section{Conclusion}

In this work, we propose the dense transformer networks to enable
the automatic learning of patch sizes and shapes in dense prediction
tasks. This is achieved by transforming the intermediate feature
maps to a different space using nonlinear transformations. A unique
challenge in dense prediction tasks is that, the spatial
correspondence between inputs and outputs should be preserved in
order to make pixel-wise predictions. To this end, we develop the
dense transformer decoder layer to restore the spatial
correspondence. The proposed dense transformer modules are
differentiable. Thus the entire network can be trained from end to
end. Experimental results show that adding the spatial transformer
and decoder layers to existing models leads to improved performance.
To the best of our knowledge, our work represents the first attempt
to enable the learning of patch size and shape in dense prediction.
The current study only adds one encoder layer and one decoder layer
in the baseline models. We will explore the possibility of adding
multiple encoder and decoder layers at different locations of the
baseline model. In this work, we develop a simple and efficient
decoder sampler for interpolation. A more complex method based on
irregular quadrilaterals might be more accurate and will be explored
in the future.

\section*{Acknowledgments}
This work was supported in part by National Science Foundation grants
IIS-1615035 and DBI-1641223, and by Washington State University. We
gratefully acknowledge the support of NVIDIA Corporation with the donation
of the Tesla K40 GPU used for this research.


{\lsp{0.85}

{\scriptsize \bibliography{DTN_nips_2017}}

\end{document}